\documentclass{article}
\usepackage[final]{corl_2018}

\usepackage[utf8]{inputenc}
\usepackage[T1]{fontenc}
\usepackage{xcolor}
\usepackage{graphicx}
\usepackage{todonotes}
\usepackage{siunitx}
\usepackage[normalem]{ulem}
\usepackage{makecell}
\usepackage{comment}
\usepackage{authblk}
\usepackage{amsmath}
\usepackage{amsfonts}
\usepackage{rotating}
\usepackage{pdflscape}
\usepackage{tabulary}
\usepackage{tabularx}
\usepackage{booktabs}
\usepackage{nicefrac}

\definecolor{green_plot}{HTML}{2CA02C}
\definecolor{cyan_plot}{HTML}{17BECF}

\newcommand*\samethanks[1][\value{footnote}]{\footnotemark[#1]}

\newcommand{\ie}{i.\,e.\ }
\newcommand{\eg}{e.\,g.\ }

\title{Modular Vehicle Control for Transferring Semantic Information Between Weather Conditions Using GANs}

\author[1,2\thanks{The majority of this work was done while at Audi Electronics Venture with equal contribution. Correspondence to: Patrick Wenzel \textless{}\texttt{wenzel@cs.tum.edu}\textgreater{}.}]{Patrick Wenzel}
\author[1,2\samethanks]{Qadeer Khan}
\author[1,2]{Daniel Cremers}
\author[1]{Laura Leal-Taix\'{e}}
\affil[1]{Technical University of Munich}
\affil[2]{Artisense}

\begin{document}
\maketitle


\begin{abstract}
Even though end-to-end supervised learning has shown promising results for sensorimotor control of self-driving cars, its performance is greatly affected by the weather conditions under which it was trained, showing poor generalization to unseen conditions. In this paper, we show how knowledge can be transferred using semantic maps to new weather conditions without the need to obtain new ground truth data. To this end, we propose to divide the task of vehicle control into two independent modules: a control module which is only trained on one weather condition for which labeled steering data is available, and a perception module which is used as an interface between new weather conditions and the fixed control module. To generate the semantic data needed to train the perception module, we propose to use a generative adversarial network (GAN)-based model to retrieve the semantic information for the new conditions in an unsupervised manner. We introduce a master-servant architecture, where the master model (semantic labels available) trains the servant model (semantic labels not available). We show that our proposed method trained with ground truth data for a single weather condition is capable of achieving similar results on the task of steering angle prediction as an end-to-end model trained with ground truth data of 15 different weather conditions.
\end{abstract}

\keywords{Imitation learning, transfer learning, modular vehicle control} 


\section{Introduction}\label{sec:intro}

One major goal of robotics and artificial intelligence research is to develop self-driving cars which can accurately perceive the environment and interact with the world. To develop an approach for addressing these problems, we have to deal with enormous challenges in perception, control, and localization. In general, the task of building an autonomous driving system can be divided into two parts: 1) path planning, and 2) vehicle control. Path planning provides a global solution for reaching a destination from a given starting position. It uses various information from different sensors such as GPS, IMU, and traffic conditions to infer the most optimized path. Meanwhile, vehicle control is meant to provide a local solution for predicting the immediate steering commands at the current instance in time. It utilizes information from sensors such as RGB cameras, lidar or radar. These sensors allow the self-driving car to sense and understand its current surroundings, such as the status of traffic lights or the presence of a pedestrian or another vehicle in front of the car.

\begin{figure}[ht]
  \centering
  \includegraphics[width=\linewidth]{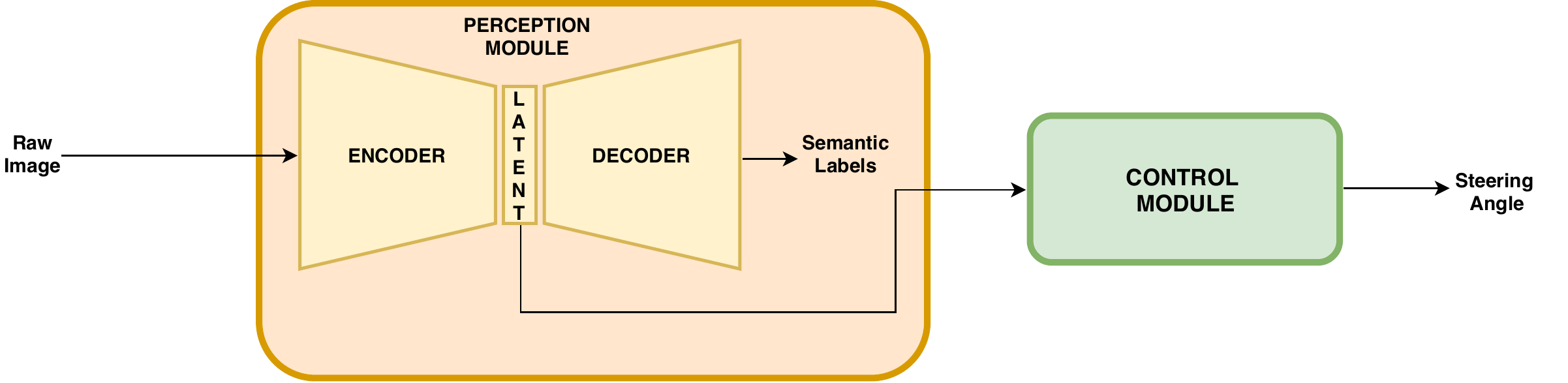}
  \caption{The perception module is trained as an encoder-decoder architecture, without any skip connections. The encoder sub-module first embeds the raw image into a lower dimensional latent vector. The decoder sub-module reconstructs the semantic scene from this latent vector. If the low dimensional latent vector contains all the necessary information to reconstruct the semantic scene to a reasonable degree of accuracy, then we directly feed it as an input to the control module instead of the semantic labels.}
  \label{fig:segcodercontrol}
\end{figure}

In this paper, we focus our attention only on vehicle control to explain how transfer learning can be utilized to improve the robustness and stability of predicting steering commands even for unseen weather conditions for which no supervised data is available. For this, the task of vehicle control is segregated into perception and control. Figure~\ref{fig:segcodercontrol} represents two modules, with each performing one of these tasks. The purpose of the perception module is to pre-process the raw input sensor data and extract useful features. In our approach, we use images captured by an RGB camera to extract semantic features of the scene. These extracted features are then fed to the control module which aims to produce the correct steering command for that particular sensor input.

{\bf Modular pipeline vs end-to-end learning.} In an end-to-end training approach, both the perception and the control module would be trained together~\citep{Chen2015}. We propose to split the task into separate perception and control, so that each module is trained and optimized independently without affecting each other. The main advantage of the separate modules is that without retraining the control module, we can simply replace the perception module to work on different environmental conditions, whereas in an end-to-end learning system, supervised labels for the new domain would first be needed to be collected and then the control module would also need to be retrained on this additional data.

Our main contributions are the following:

\begin{itemize}
  \item Ability to control the vehicle in unseen weather conditions without having the need to collect additional data for the steering commands and without requiring to retrain the control module. This is done by simply replacing the perception module additionally trained on the semantics of the new condition.
  \item We show how knowledge can be transferred from a weather condition for which semantic labels are available to other weather conditions for which no labels exist in an unsupervised manner using GANs.
\end{itemize}

\section{Related Work}\label{sec:background}

{\bf Supervised learning for self-driving cars.} The use of supervised learning methods to train driving policies for self-driving cars is a well-known and common approach. The first step towards using neural networks for the task of road following dates back to ALVINN~\citep{Pomerleau1988}. This approach uses a very simple shallow network which maps images and a laser range finder as input and produces action predictions. Recently, NVIDIA~\citep{Bojarski2016} proposed to use deep convolutional neural networks trained end-to-end for a simple lane following task. This approach was successful in relatively simple real-world scenarios. One major drawback of end-to-end imitation learning is that it cannot generalize well across different domains for which no labeled training data is available. However, most end-to-end learning approaches~\citep{Zhang2016d,Silver2010,LeCun2006} suffer from this problem.

{\bf Transfer learning.} Generative adversarial networks provide a framework to tackle this generalization gap~\citep{Karacan2016} by image generation techniques which can be used for domain adaptation. The authors of~\citep{You2017} proposed a network that can convert non-realistic virtual images into realistic ones with similar scene structure. Similarly, Hoffman et al.~\citep{Hoffman2017} proposed a novel discriminatively-trained adversarial model which can be used for domain adaptation in unseen environments. They show new state-of-the-art results across multiple adaptation tasks, including digit classification and semantic segmentation of road scenes.

{\bf Semantic segmentation.} Visual understanding of complex environments is an enabling factor for self-driving cars. The authors of~\citep{Cordts2016} provide a large-scale dataset with semantic abstractions of real-world urban scenes focusing on autonomous driving. By using semantic segmentation, it is possible to decompose the scene into pixel-wise labels we are particularly interested in. This especially helps self-driving cars to discover driveable areas of the scene. It is therefore possible to segment a scene into different classes (\eg road and not road)~\citep{Chen2017a}. 

{\bf Modular pipeline vs end-to-end learning.} The authors of~\citep{Dosovitskiy2017} trained both an end-to-end and a modular based model on one set of weather conditions and tested the model on a different set of weather conditions. Based on their experiments they concluded that the modular approach is more vulnerable to failures under complex weather conditions than the end-to-end approach.

Our method also uses a modular approach, but additionally introduces an image translation technique to overcome the generalization gap between the unseen weather conditions. This is done by only retraining the perception module without having to retrain the control module for each and every domain (\ie weather condition). A useful consequence of this is that we do not have to recollect additional labeled data for the new conditions.

\section{Imitation Learning on the Latent Semantic Vector}\label{sec:imitation-learning}

\begin{figure}[ht]
  \centering
  \includegraphics[width=0.8\linewidth]{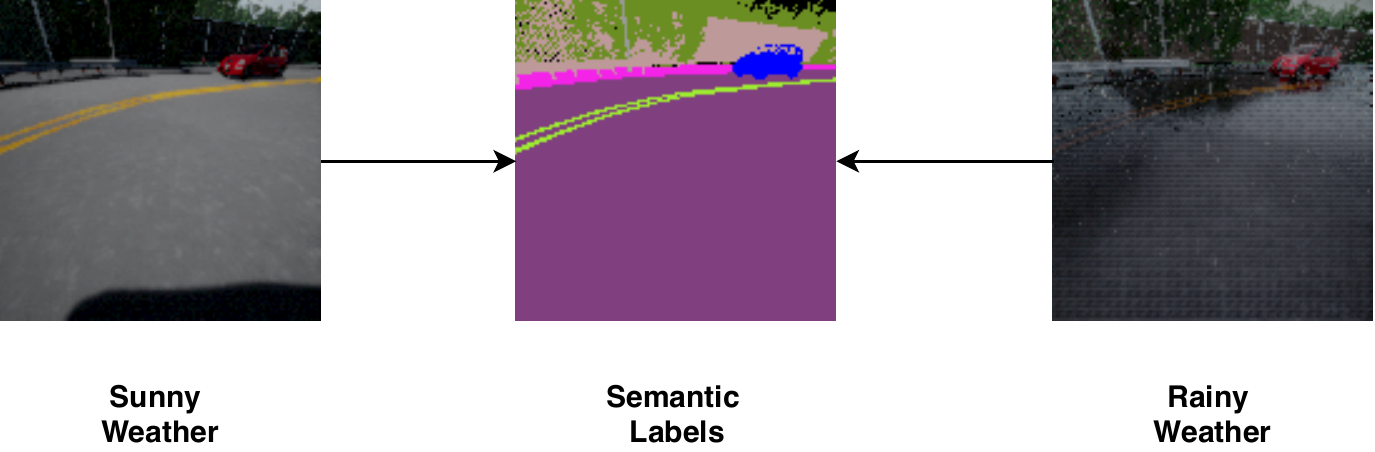}
  \caption{For the perception module we take in raw image data as obtained from the car's camera and output the semantic segmentation of the scene. Notice that irrespective of the weather condition the semantics of the scene remain the same. Since the perception module bears the burden of producing the correct semantic labels, the control module would be robust to changes in lighting, weather, and climate conditions.}
  \label{fig:whysegment}
\end{figure}

{\bf Perception module.} In this work, we use images captured by an RGB camera placed at the front of the car as inputs to the perception module. The perception module processes these images and produces an output map containing the semantics of the scene, which in turn can be used as an input to the control module. The CARLA~\citep{Dosovitskiy2017} simulator yields semantic labels for 13 classes. The advantage of using semantic labels instead of raw RGB data is described below:

\begin{itemize}
  \item Figure~\ref{fig:whysegment} shows how two weather conditions have different RGB inputs but the same semantic pixel labels. Hence, the control module does not separately need to learn to predict the correct steering commands for each and every weather condition. 
  \item The semantic labels can precisely localize the pixels of important road landmarks such as traffic lights and signs. The status/information contained on these can then be read off to take appropriate planning and control decisions.
  \item A high proportion of the pixels have the same label as its neighbors. This redundancy can be utilized to reduce the dimensionality of the semantic scene. Hence, the number of parameters required to train the control module can then also be reduced. 
\end{itemize}

The perception module, which is used to produce the semantic labels of a scene from the RGB camera is trained as an encoder-decoder architecture. The network architecture which is being used is a modified version of the one proposed by Larsen et al.~\citep{Larsen2015}. The structure and the parameters of the model are shown in the supplementary material. The encoder first encodes the information contained in the input data to a lower dimensional latent vector. The decoder, then takes this latent vector and attempts to reconstruct the semantics of the scene. The output of the decoder is of the same size as the image but having 13 channels with each representing the probability of occurrence of one of the semantic labels. The model is trained by minimizing the weighted sum of the categorical cross-entropy of each pixel in the image. The categorical cross-entropy (negative log-likelihood) between predictions $p$ and targets $t$ is calculated as follows:

\begin{equation*}
  \mathcal{L}_i = - \sum_j \, t_{i,j} \log(p_{i,j}) w_j,
\end{equation*}

where $i$ denotes the pixel and $j$ denotes the class. The weight $w_j$ of each semantic label is inversely proportional to its frequency of occurrence in the dataset.

{\bf Control module.} Note that we do not use skip connections between the encoder and decoder of the perception module. Therefore, since the lower dimensional latent vector is capable of reconstructing the semantic labels of the scene, we can directly use this vector as input to the control module instead of the complete scene. Figure~\ref{fig:segcodercontrol} depicts how the latent semantic embedding vector produced by the encoder of the perception module can be used as an input to the control module.

The control module aims to predict the correct steering angle, from the latent embedding fed as an input to the model. The data used for training the control module is collected in a supervised manner by recording images and their corresponding steering angles. The loss function attempts to minimize the mean squared error (MSE) between the actual steering angle and the one predicted by the model across all the samples. The architecture of the control model is depicted in the supplementary material.

\section{Master-Servant Architecture for Transfer Learning}\label{sec:transfer-learn}

\begin{figure}[ht]
  \centering
  \includegraphics[width=0.9\linewidth]{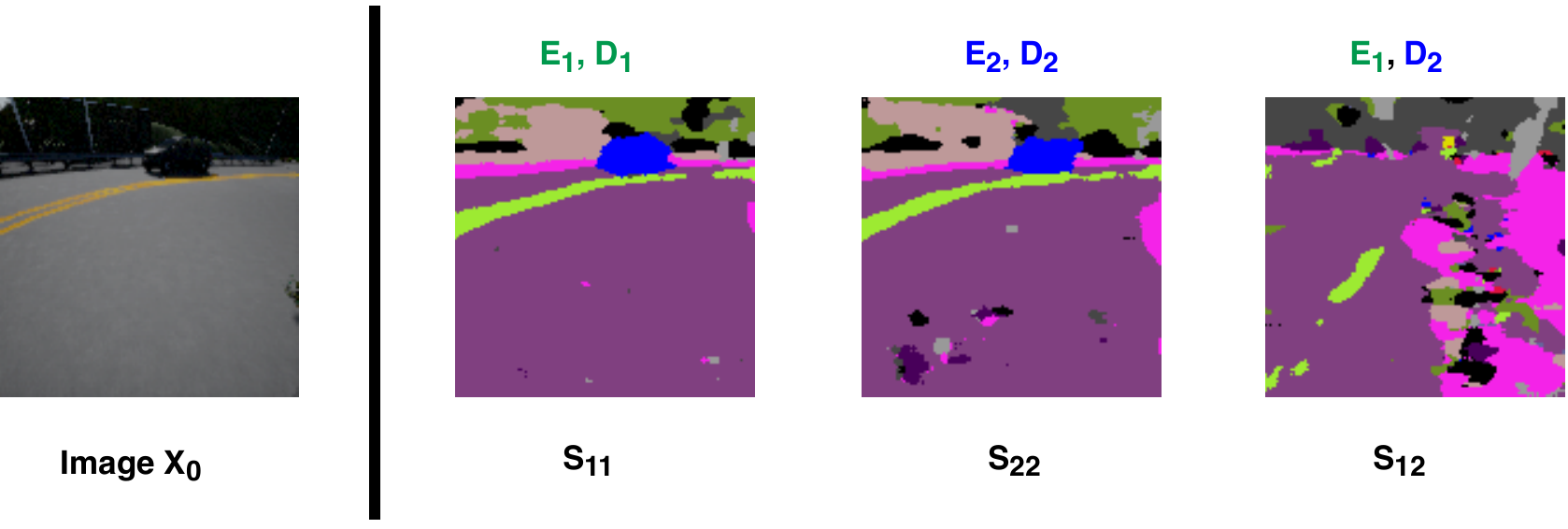}
  \caption{This figure shows the segmentation reconstructions $S_{11}$ and $S_{22}$ when image $X_0$ is passed through two segmentation models $M_1$ (with Encoder $E_1$, Decoder $D_1$) and $M_2$ (with Encoder $E_2$, Decoder $D_2$). Both models are trained independently on the same data. Note that the reconstructions reflect the true semantics of the scene reasonably well. $S_{12}$ shows the reconstruction when the embedding from encoder $E_1$ is fed to through decoder of $D_2$. The ambiguity in $S_{12}$ implies that for the same image the two models yield different semantic vectors.}
  \label{fig:diffSemantics}
\end{figure}

The control module does not perform well if tested in an environment which is completely different from the one on which the perception module was trained on. A naive and yet computational demanding solution could be to retrain the perception module under every other weather condition. However, this is not a viable solution for the following reasons:

\begin{itemize}
  \item We would need semantic labels for every other weather condition. Obtaining semantic labels of a scene is a painstakingly slow process and prone to errors, since it requires human effort. 
  \item Even if we have access to the semantic labels and retrain the perception module under the new environmental conditions, we would still have to also retrain the control module. This is due to the fact that the semantic latent vector produced by the new perception module might be different from the one produced by the old perception module, despite the same semantics of the scene. Figure~\ref{fig:diffSemantics} describes how for the same image, two independently trained segmentation models could yield different semantic vectors, despite being trained on the same data. 
\end{itemize}

\begin{figure}[ht]
  \centering
  \includegraphics[width=0.95\linewidth]{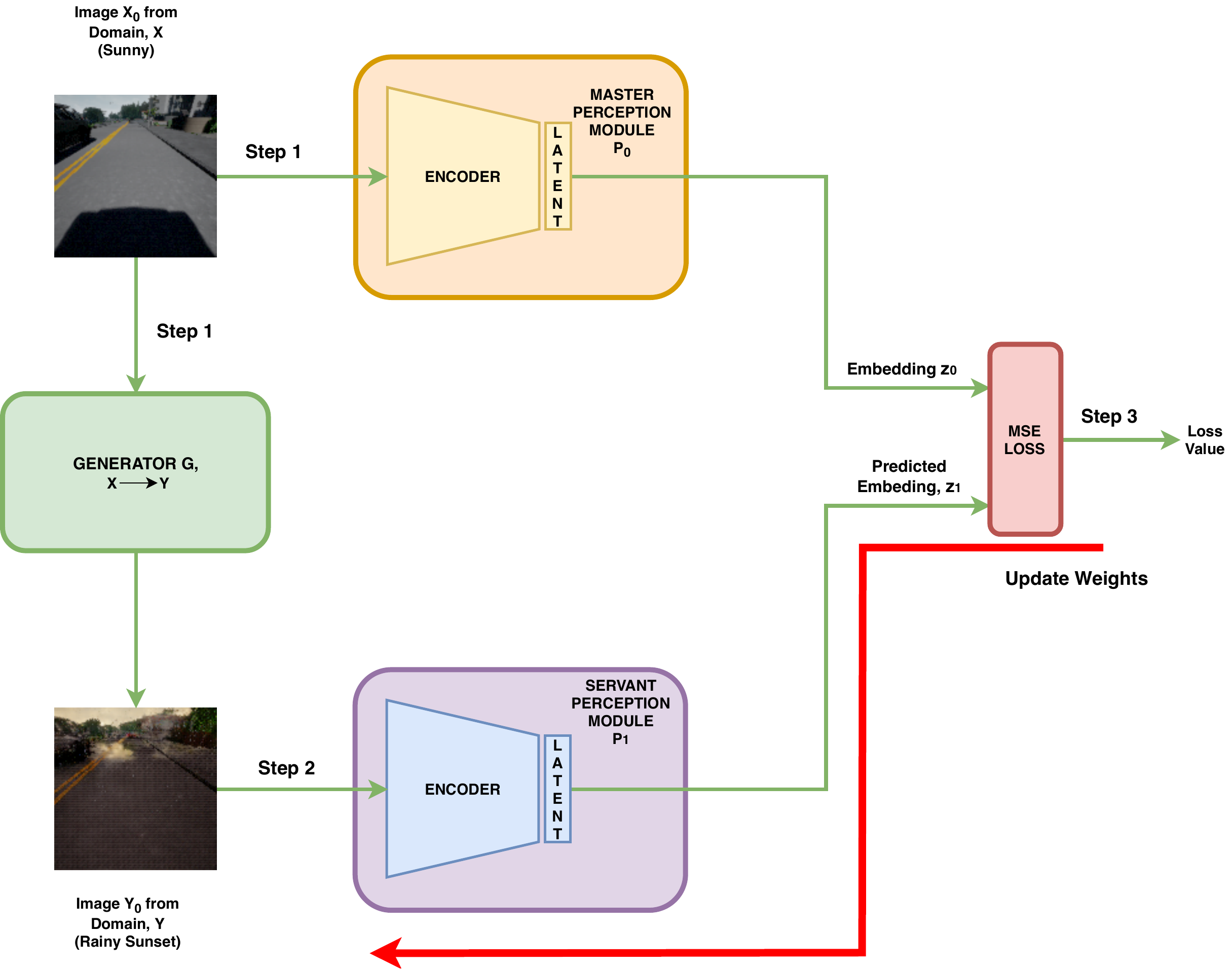}
  \caption{We propose a master-servant architecture to train a servant perception module $P_1$ for images which do not have semantic labels in an unsupervised manner. Images in domain $X$ have semantic labels and are used to train the perception module $P_0$, which we refer to as the master perception module. $P_0$ is pre-trained using the complete encoder-decoder architecture. Images in domain $Y$ do not have semantic labels. The process works as follows. \textbf{Step~1:} The generator $G$ is used to convert an image $X_0$ from domain $X$ to an image $Y_0$ in domain $Y$ such that the semantic information is preserved. Meanwhile $X_0$ is also fed to the master perception module $P_0$ to get the latent embedding $z_0$. \textbf{Step~2:} The image $Y_0$ is fed to the servant perception module $P_1$ to get the predicted latent embedding $z_1$. \textbf{Step~3:} Since the semantic labels of $X_0$ and $Y_0$ are the same, their latent embeddings should also be the same. We use the mean squared error (MSE) to minimize this difference, wherein the embedding $z_0$ is used as the true label for training $P_1$. \textbf{Update Weights:} We back-propagate the MSE loss to update the weights of only $P_1$ so that its embedding matches with that of $P_0$. The green arrows indicate forward propagation and the red arrow shows back-propagation.}
  \label{fig:masterservant}
\end{figure}

{\bf Proposed master-servant architecture.} Suppose that the perception module $P_0$ and the control module $C_0$ are trained under a certain environmental condition. When tested on a very different weather condition $P_0$ may fail to produce the relevant semantic latent vector for the control module $C_0$ to take the correct steering decision. We would therefore like to replace $P_0$ with a different perception module $P_1$ such that it produces the correct latent vector to allow the same control module $C_0$ to execute the appropriate steering command even on this very different condition. For this, we propose a master-servant architecture model for training the perception module functioning on images from a domain for which no semantic labels are available. Figure~\ref{fig:masterservant} demonstrates the necessary steps of the master-servant architecture.

Suppose we have images (from domain $X$) and their corresponding semantic labels. With this, we can train a segmentation model using the encoder-decoder architecture described previously. We refer to the trained encoder of this model as the \textbf{master perception module} $P_0$. We would also like to obtain the correct semantic embedding of images (from domain $Y$) for new conditions for which no semantic labels are available. We refer to the perception module for which we would like to furnish the correct semantic embedding for images in domain $Y$ as the \textbf{servant perception module} $P_1$. We use the master module, $P_0$, to train the servant module, $P_1$, in the steps described as follows:

\begin{enumerate}
  \item An image $X_0$ is arbitrarily selected from domain $X$. $X_0$ is fed to through $P_0$ to obtain the semantic embedding of the scene denoted by $z_0$. Meanwhile, the generator $G$ translates the image $X_0$ to generate an image $Y_0$ from domain $Y$, such that the semantics of the scene are preserved. If semantics are being preserved, then $z_0$ should be equal to $z_1$ (the semantic embedding obtained by feeding $Y_0$ through $P_1$).
  \item  $Y_0$ is fed through $P_1$ to get the predicted latent embedding $z_1$.
  \item The mean squared error (MSE) between $z_0$ and $z_1$ is used as the loss function to update the weights of $P_1$ in order to minimize the difference between the two latent embeddings. 
\end{enumerate}

Some examples of the images produced by the generator $G$, segmentation reconstruction when $z_0$ (semantic embedding of the master) and $z_1$ (semantic embedding of the servant) is fed through the decoder of the master perception module $P_0$ are shown in the supplementary material. 

{\bf Unsupervised transfer of semantics.} We observe that with this master-servant architecture we are able to train the servant perception module for obtaining the correct semantic embeddings for images from domain $Y$ for which semantic labels were never available. We can thus replace $P_0$ with $P_1$ which would also work on these unseen weather conditions without having to retrain the control module. Moreover, no additional human effort is required for the labeling of semantics.

The most critical component which made the functioning of this approach possible is the generator $G$, which is able to translate images between two different domains, while preserving the semantics. The generator $G$ is pre-trained using the CycleGAN~\citep{Zhu2017a} approach. Unlike other image-to-image translation methods such as pix2pix~\citep{Isola2017}, an important feature of CycleGANs is the fact that this approach does not require paired data between two domains. Therefore, the task of collecting (if even possible) images with a one-to-one correspondence between two domains can be eliminated. The procedure for training the generator $G$ using the CycleGAN approach is shown in the supplementary. The architecture used was taken from~\citep{Zhu2017a}. The supplementary material shows some examples of paired and unpaired data from two different domains produced by the CARLA simulator.

\section{Experimental Results}\label{sec:experiments-results}

{\bf Experimental setup.} For evaluating our method, we used the CARLA simulator. The CARLA simulator provides 15 different weather conditions (labeled from 0 to 14). We focus our attention on the car turning around corner scenarios since it is a more complicated maneuver to perform than lane following and it would thus give a better understanding of possible failure conditions. We train 5 different models to predict the steering angle whilst assuming that the car throttle is fixed. For a fair comparison, the approach is evaluated on multiple different turns and we do not consider the presence of pedestrians and cars in the ego vehicle's driving lane. The starting position of the agent is just before the curve and the duration of the turn is fixed to 120 frames since it covers the entire turning maneuver. The turn is considered successful if the car did not crash whilst executing the turn. Furthermore, in order to make a quantitative evaluation of the performance of each of the 5 models, new test data containing the images and the corresponding true steering commands for each of the 15 weather conditions was collected. Figure~\ref{fig:modelsComp} shows a plot of the mean squared error (MSE) between the actual and the predicted steering commands by the 5 different models across all the weather conditions on samples of the test data. Meanwhile, Table~\ref{tab:turns} enumerates the percentage of turns each of the 5 models are successfully able to execute across all the 15 weather conditions.

The supplementary material contains the description and some samples of the 15 weather conditions along with video samples demonstrating the performance of the models on certain weather conditions. The dataset can be downloaded at: \url{https://git.io/fApfH}. The details of the 5 models are given below:

\begin{figure}[ht]
  \centering
  \includegraphics[width=0.85\linewidth]{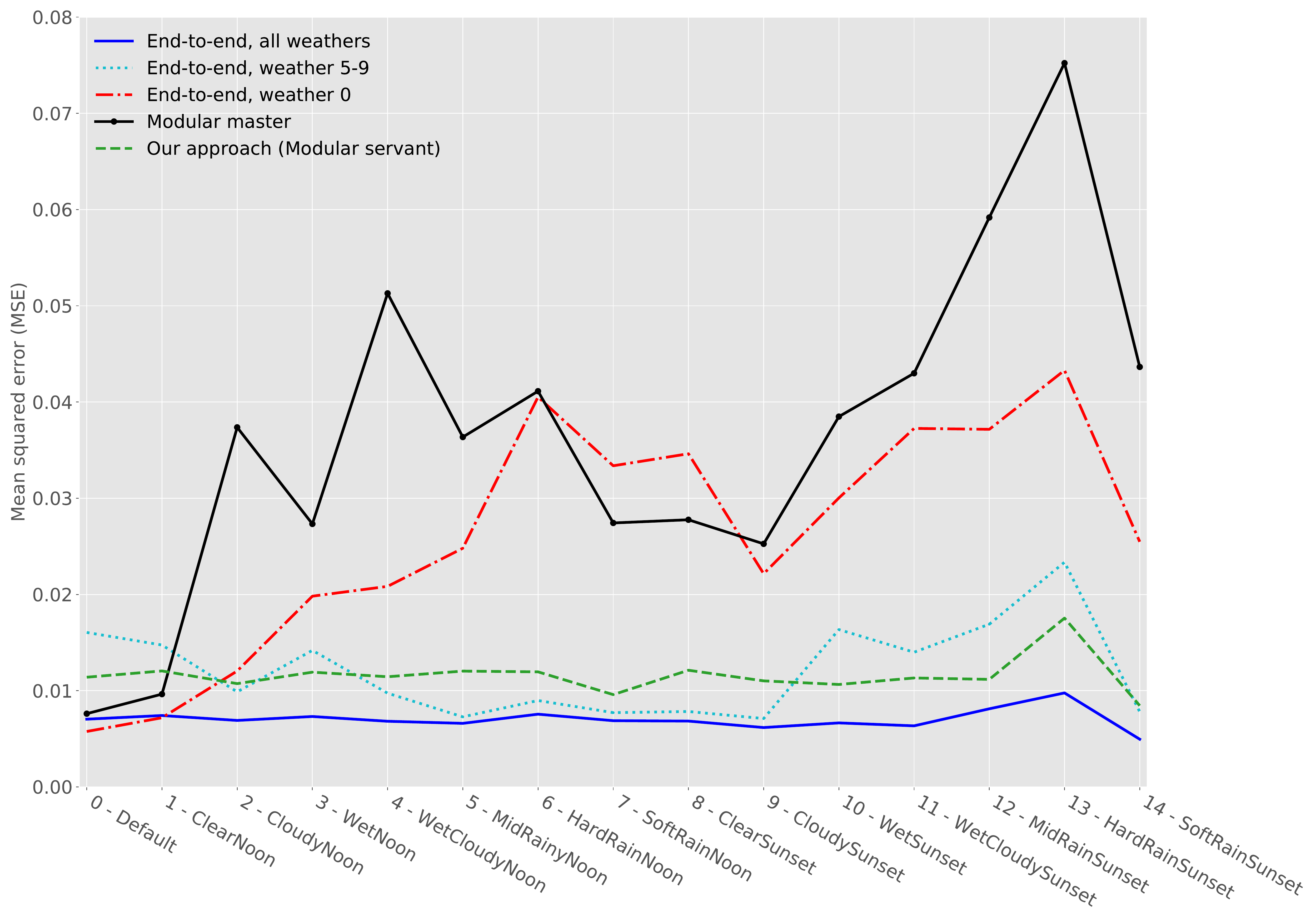}
  \caption{Plot of the mean squared error (MSE) between the actual and the predicted steering commands by 5 different models across the weather conditions 0 to 14. The \textbf{\textcolor{blue}{blue}} line is the error plot for a model trained end-to-end, from images and corresponding steering commands for all the 15 weather conditions. The \textbf{\textcolor{cyan_plot}{cyan}} error curve corresponds to the end-to-end model trained on images and steering commands for weathers 5-9. The \textbf{\textcolor{red}{red}} line is for the model trained end-to-end from images and corresponding steering commands for only the default weather condition 0. The \textbf{\textcolor{black}{black}} line represents the model referred to as the master whose perception and control modules are trained separately. The perception module is trained using the actual semantic labels available for the default weather condition, whereas the control model is trained from the actual steering commands of the same condition. The \textbf{\textcolor{green_plot}{green}} curve is the model whose control model is the same as the one for the master, but the perception module is trained as a servant from the master perception module from images generated by the CycleGANs for weather conditions 2, 3, 4, 6, 8, 9, 10, 11, 12, and 13, in addition to the default condition 0.}
  \label{fig:modelsComp}
\end{figure}

{\bf End-to-end, all weathers.} An end-to-end model is trained on all weather conditions. Here we have assumed that we have access to the steering commands across all the conditions. As can be seen from Figure~\ref{fig:modelsComp}, this model gives the lowest error particularly for weathers 1 to 14. Moreover, we observe in Table~\ref{tab:turns} that this model is able to successfully execute a high proportion of the turns across all the weather conditions, since it was trained on all of them. All subsequent models are trained with the steering commands available for a subset of the weather conditions and their performance is compared with this model.

{\bf End-to-end, weather 5-9.} This model is trained end-to-end on weathers 5, 6, 7, 8, and 9 which were arbitrarily selected just to see how it would perform on unseen weather conditions. As shown in Figure~\ref{fig:modelsComp} it has a relatively low error on these conditions and a higher error elsewhere. Furthermore, the plot shows that this end-to-end approach only seems to work well on the trained conditions for which we have labeled data. Moreover, as can be seen in Table~\ref{tab:turns}, the model is capable of maneuvering well on the trained weather conditions and on those which are similar or have good visibility. However, on weather conditions 11-14 the model fails to execute the majority of the turns. This is mainly due to the fact that these weather conditions (11-14) are relatively disparate in terms of appearance and visibility as compared to the trained ones (5-9). 

In practice, we do not have the steering commands available for all the possible or even a diverse subset of the weather conditions. Rather, the labeled data would correspond to only the condition of the day/period on which it was collected. Therefore, the 3 successive models that we now consider assume that the steering commands and the corresponding images/semantics are only available for the default weather condition (labeled as 0). From this, we evaluate how end-to-end training would compare to the proposed modular approach across all the remaining weather conditions for which no labeled data is available. 

{\bf End-to-end, weather 0.} This model is trained end-to-end from images and steering commands for the default weather condition. Figure~\ref{fig:modelsComp}, shows that this model outperforms all the other models only on weather condition 0 on which it was trained. For all other conditions, it gives high errors.

{\bf Modular master.} This model is trained on the default weather condition (0) but the task is divided into 2 separate perception and control modules. The perception module $P_0$ is trained on the semantic labels. We refer to this perception module as the master, since it will later be used to train the servant module for retrieving the semantic information of the unseen weather conditions. The control module is in turn trained with imitation learning to predict the steering angle of the car from the latent embedding generated by the encoder of $P_0$. The forth row of Table~\ref{tab:turns} depicts the percentage of turns the model was successfully able to maneuver for each of the 15 conditions. As observed in the table, the model is successful only on the default weather conditions (on which it was trained) and the sunny weather condition (which closely resembles the default condition). Similar to the previous model (trained end-to-end on the default condition), this model also fails on a large proportion when tested on weather conditions that are far off from the default condition in terms of visual appearance. From this, there seems to be no apparent advantage of using a modular approach over the end-to-end training when we have access to the labels for only one weather condition. Nevertheless, the master perception module $P_0$ obtained through this method will serve as a baseline for training a servant perception module that additionally works for unseen weather conditions. This approach is described in the following. 

{\bf Our approach (Modular servant).} We train one servant perception module to cater for weather conditions on which $P_0$ failed to perform. We selected a subset of weather conditions (\ie 2, 3, 4, 6, 8, 9, 10, 11, 12, and 13) to train the servant module. Using CycleGANs, separate generators were trained between each of these conditions and the default weather condition. The images produced by the CycleGAN generators for each of these conditions were fed as an input in equal proportion along with the default images to train only a single servant perception module $P_1$. Despite having no access to the steering commands and the semantic labels for weather conditions 1 to 14, Figure~\ref{fig:modelsComp} shows that the error for this model across these 14 weather conditions is significantly lower than the previous 2 models which were also trained only from labels of weather condition 0. Moreover, we see from the last row of Table~\ref{tab:turns}, that this model is successfully able to execute a good proportion of the turns for most of the weather conditions. Only on condition 13 (HardRainSunset), the model fails to perform well. The visibility under this condition is low and the images generated by the CycleGAN do not seem to preserve the semantics, hence resulting in the model to perform relatively poorly. Nevertheless, on all the other remaining weather conditions its performance is comparable to the first end-to-end model trained on steering labels for all the weather conditions.

\begin{table}[ht]
    \centering
    \caption{The table reports the percentage of successfully completed turns by the 5 models for each weather condition. Higher is better.}
    \resizebox{\linewidth}{!}{  
    \begin{tabular}{lccccccccccccccc|c}
        \toprule
        & \multicolumn{15}{c|}{Weather condition} & \\
        Model & 0 & 1 & 2 & 3 & 4 & 5 & 6 & 7 & 8 & 9 & 10 & 11 & 12 & 13 & 14 & average \\
        \midrule
        End-to-end, all weathers & 100 & 100 & 88 & 88 & 100 & 100 & 100 & 100 & 100 & 100 & 100 & 88 & 88 & 88 & 100 & 96 \\
        End-to-end, weather 5-9 & 88 & 88 & 100 & 88 & 100 & 100 & 100 & 100 & 100 & 100 & 100 & 50 & 50 & 25 & 88 & 85 \\
        End-to-end, weather 0 & 100 & 63 & 38 & 38 & 13 & 13 & 0 & 0 & 0 & 50 & 13 & 0 & 0 & 0 & 0 & 22 \\
        Modular master & 100 & 100 & 88 & 50 & 50 & 63 & 50 & 50 & 50 & 63 & 75 & 50 & 0 & 0 & 50 & 56 \\
        \bf{Our approach (Modular servant)} & 100 & 100 & 100 & 100 & 88 & 100 & 100 & 100 & 100 & 100 & 100 & 88 & 100 & 63 & 100 & 96 \\
        \bottomrule
    \end{tabular}
    \label{tab:turns}
    }
\end{table}

\section{Conclusion}\label{sec:conclusion}

In this paper, we have shown that in order to generalize vehicle control across unseen weather conditions it is worthwhile to divide the task into separate perception and control modules. This separation eliminates the tedious task of recollecting labeled steering command data for each and every new environment the vehicle might come across. Moreover, retraining of the control module for new environments can be avoided by a simple replacement of the perception module. The initial perception module was trained from the semantic labels available only for one of the weather conditions. For environments for which semantic labels are missing, the proposed master-servant architecture can be deployed for transferring semantic knowledge from one domain to another (\ie between different weather conditions) in an unsupervised manner using CycleGANs which do not require paired data. We believe that the presented approach to making driving policies more robust by training under different weather conditions will prove useful in future research.



\clearpage
\acknowledgments{This research was partially funded by the Humboldt Foundation through the Sofja Kovalevskaja Award.}


\bibliography{main}  

\newpage

\renewcommand{\thesection}{S.\arabic{section}}   
\renewcommand{\thetable}{S.\arabic{table}}   
\renewcommand{\thefigure}{S.\arabic{figure}}
\setcounter{section}{0}
\setcounter{figure}{0}
\setcounter{table}{0}

\section*{Supplementary Material}\label{part:supplementary}

\begin{table}[ht]
    \centering
    \caption{Encoder-decoder architecture used to train the segmentation perception module for the master and all servant models. The convolution layers numbered 15 and 16 have a kernel size of 4, stride of 1 and no padding. All other convolution layers have kernel size 4, stride of 2 and padding of 1. All the Leaky ReLU activation functions have a negative slope of \num{0.2}. The output of the model has 13 channels with each corresponding to one of the semantic labels. The output of the last layer of the encoder (Layer 15) is fed to the control module to predict the correct steering direction. The same layer is also used to train the encoders of all servant modules. The code and model of the architecture is a modified version of~\url{https://github.com/seangal/dcgan_vae_pytorch}.}
    \resizebox{\linewidth}{!}{
    \vspace{.2cm}
    \begin{tabular}{cccc|cccc}
        \toprule
        \multicolumn{4}{c|}{\bf{\textsc{Encoder}}} & \multicolumn{4}{c}{\bf{\textsc{Decoder}}} \\
        \midrule
        \bf{Layer Number} & \bf{Layer Type} & \bf{Layer Input} & \bf{Layer Output} & \bf{Layer Number} & \bf{Layer Type} & \bf{Layer Input} & \bf{Layer Output} \\
        \midrule
        1 & Convolution & $3 \times 128 \times 128$ & $32 \times 64 \times 64$ & 16 & Convolution (Transpose) & $64 \times 1 \times 1$ & $512 \times 4 \times 4$ \\
        2 & Leaky ReLU activation & $32 \times 64 \times 64$ & $32 \times 64 \times 64$ & 17 & Batch normalization & $512 \times 4 \times 4$ & $512 \times 4 \times 4$ \\
        3 & Convolution & $32 \times 64 \times 64$ & $64 \times 32 \times 32$ & 18 & Leaky ReLU activation & $512 \times 4 \times 4$ & $512 \times 4 \times 4$ \\
        4 & Batch normalization & $64 \times 32 \times 32$ & $64 \times 32 \times 32$ & 19 & Convolution (Transpose) & $512 \times 4 \times 4$ & $256 \times 8 \times 8$ \\
        5 & Leaky ReLU activation & $64 \times 32 \times 32$ & $64 \times 32 \times 32$ & 20 & Batch normalization & $256 \times 8 \times 8$ & $256 \times 8 \times 8$ \\
        6 & Convolution & $64 \times 32 \times 32$ & $128 \times 16 \times 16$ & 21 & Leaky ReLU activation & $256 \times 8 \times 8$ & $256 \times 8 \times 8$ \\
        7 & Batch normalization & $128 \times 16 \times 16$ & $128 \times 16 \times 16$ & 22 & Convolution (Transpose) & $256 \times 8 \times 8$ & $128 \times 16 \times 16$ \\
        8 & Leaky ReLU activation & $128 \times 16 \times 16$ & $128 \times 16 \times 16$ & 23 & Batch normalization & $128 \times 16 \times 16$ & $128 \times 16 \times 16$ \\
        9 & Convolution & $128 \times 16 \times 16$ & $256 \times 8 \times 8$ & 24 & Leaky ReLU activation & $128 \times 16 \times 16$ & $128 \times 16 \times 16$ \\
        10 & Batch normalization & $256 \times 8 \times 8$ & $256 \times 8 \times 8$ & 25 & Convolution (Transpose) & $128 \times 16 \times 16$ & $64 \times 32 \times 32$ \\
        11 & Leaky ReLU activation & $256 \times 8 \times 8$ & $256 \times 8 \times 8$ & 26 & Batch normalization & $64 \times 32 \times 32$ & $64 \times 32 \times 32$ \\
        12 & Convolution & $256 \times 8 \times 8$ & $512 \times 4 \times 4$ & 27 & Leaky ReLU activation & $64 \times 32 \times 32$ & $64 \times 32 \times 32$ \\
        13 & Batch normalization & $512 \times 4 \times 4$ & $512 \times 4 \times 4$ & 28 & Convolution (Transpose) & $64 \times 32 \times 32$ & $32 \times 64 \times 64$ \\
        14 & Leaky ReLU activation & $512 \times 4 \times 4$ & $512 \times 4 \times 4$ & 29 & Batch normalization & $32 \times 64 \times 64$ & $32 \times 64 \times 64$ \\
        15 & Convolution & $512 \times 4 \times 4$ & $64 \times 1 \times 1$ & 30 & Leaky ReLU activation & $32 \times 64 \times 64$ & $32 \times 64 \times 64$ \\
        & & & & 31 & Convolution (Transpose) & $32 \times 64 \times 64$ & $13 \times 128 \times 128$ \\
        & & & & 32 & Sigmoid activation & $13 \times 128 \times 128$ & $13 \times 128 \times 128$ \\
        \bottomrule
    \end{tabular}
    \label{tab:segcoderArchi}
    }
\end{table}

\begin{table}[ht]
    \centering
    \caption{Architecture of the control model. Note that the input to the control module is a vector of size 64, corresponding to the size of the latent embedding produced by the encoder of the perception module.}
    \begin{tabular}{cccc}
        \toprule
        \bf{Layer Number} & \bf{Layer Type} & \bf{Layer Input} & \bf{Layer Output} \\
        \midrule
        1 & Fully connected & 64 & 100 \\
        2 & ReLU activation & 100 & 100 \\
        3 & Fully connected & 100 & 50 \\
        4 & ReLU activation & 50 & 50 \\
        5 & Fully connected & 50 & 25 \\
        6 & ReLU activation & 25 & 25 \\
        7 & Fully connected & 25 & 15 \\
        8 & ReLU activation & 15 & 15 \\
        9 & Fully connected & 15 & 8 \\
        10 & ReLU activation & 8 & 8 \\
        11 & Fully connected & 8 & 1 \\
        \bottomrule
    \end{tabular}
    \label{tab:controlArchi}
\end{table}

\begin{figure}[ht]
  \centering
  \includegraphics[width=\linewidth]{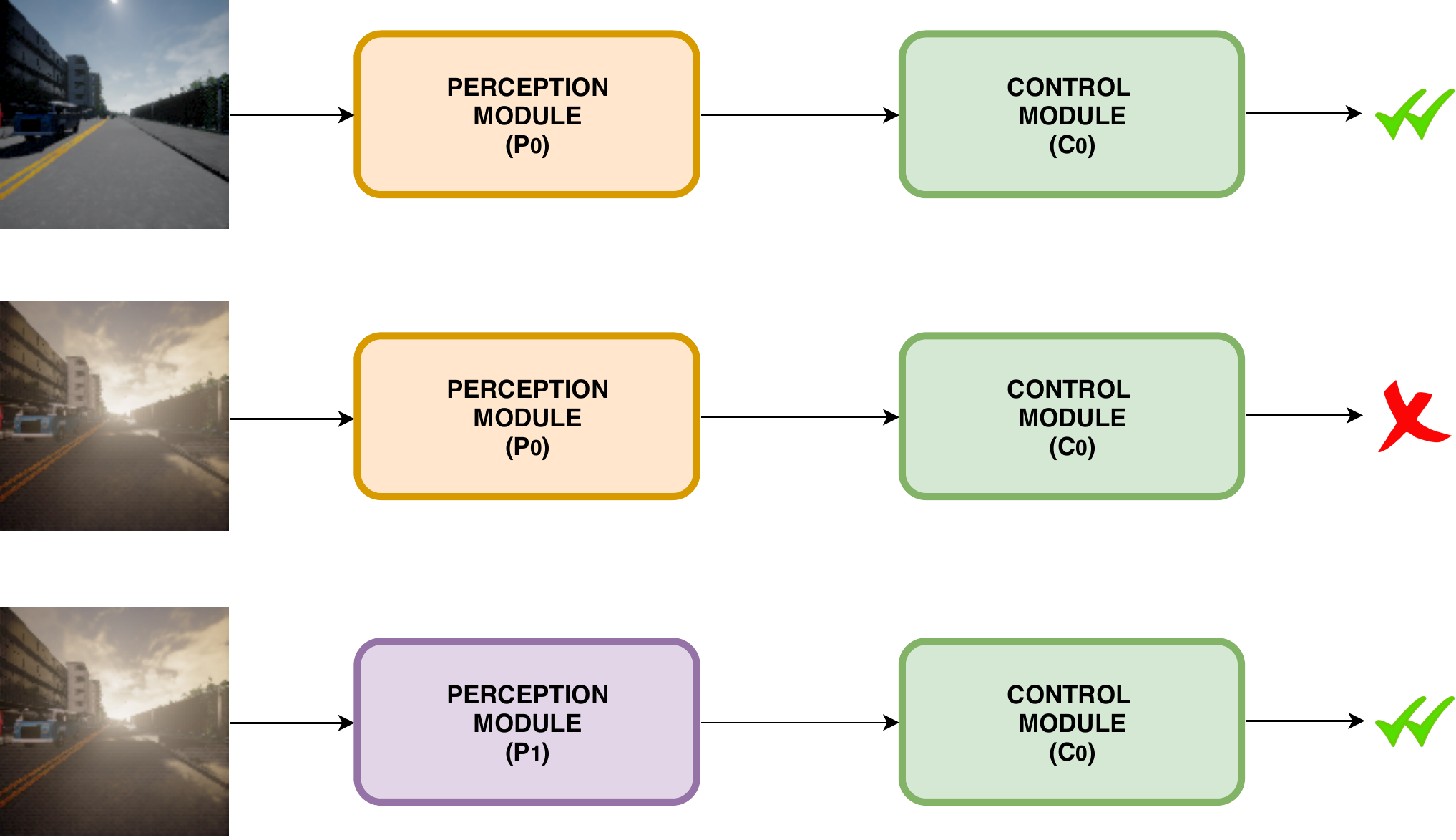}
  \caption{To obtain a good segmentation based perception module, semantic labels for a diverse range of environmental conditions are required. This may not always be the case since semantic labeling is a tedious and error-prone process. Hence, we may have only access to a limited subset of the labeled data. \textbf{Top:} This figure shows a perception module $P_0$ trained only on sunny weather conditions. Hence, when a similar data is fed to $P_0$ at test time the control model $C_0$ performs as per expectation. \textbf{Center:} This figure demonstrates that if data from a different weather condition is fed to $P_0$, the control module $C_0$ may not necessarily perform as desired. \textbf{Bottom:} This figure shows that we would like to replace $P_0$ with $P_1$, such that $P_1$ is capable of handling this unseen environment in a manner to retain the same semantic embedding. Hence, we can use the same control module $C_0$ with $P_1$.}
  \label{fig:whycgans}
\end{figure}

\begin{sidewaystable}[ht]
    \centering
    \caption{Table describing the contents and performance of the videos. Note that for the modular approach the control module was only trained on weather 0. Video pairs (5/6, 7/8, and 9/10) have the same starting position to compare between the modular master and our proposed modular servant approach. The videos are available at \url{https://www.youtube.com/playlist?list=PLbT2smuiIncsR_s9YA6KFpsa8gMwus5u7}.} 
    \vspace{.2cm}
    \begin{tabular}{llcl}
        \toprule
        \bf{Video Id} & \bf{Model} & \bf{Tested on Weather} & \bf{Comments} \\
        \midrule
        video1 & End-to-end, weather 0 & 0 & successfully turning \\
        \midrule
        video2 & Modular master & 0 & successfully turning \\   
        \midrule
        video3 & End-to-end, weather 5-9 & 8 & successfully turning \\
        \midrule
        video4 & End-to-end, weather 5-9 & 3 & crashing, model only trained on weather 5-9 \\
        \midrule
        video5 & Modular master & 2 & crashing, segmentation module only trained on weather 0 \\
        \midrule
        video6 & Our approach (Modular servant) & 2 & successfully turning, segmentation module trained with master-servant architecture \\
        \midrule
        video7 & Modular master & 4 & crashing, segmentation module only trained on weather 0 \\
        \midrule
        video8 & Our approach (Modular servant) & 4 & successfully turning, segmentation module trained with master-servant architecture \\
        \midrule
        video9 & Modular master & 11 & crashing, segmentation module only trained on weather 0 \\
        \midrule
        video10 & Our approach (Modular servant) & 11 & successfully turning, segmentation module trained with master-servant architecture \\
        \midrule
        video11 & End-to-end, all weathers & 13 & successfully turning \\
        \midrule
        video12 & End-to-end, weather 0 & 13 & crashing, model only trained on weather 0 \\
        \midrule
        video13 & Modular master & 13 & crashing, segmentation module only trained on weather 0 \\
        \midrule
        video14 & Our approach (Modular servant) & 13 & turning, unstable segmentation \\
        \midrule
        video15 & Our approach (Modular servant) & 13 & crashing, unstable segmentation \\
        \bottomrule
    \end{tabular}
\end{sidewaystable}

\begin{figure}[ht]
  \centering
  \includegraphics[width=\linewidth]{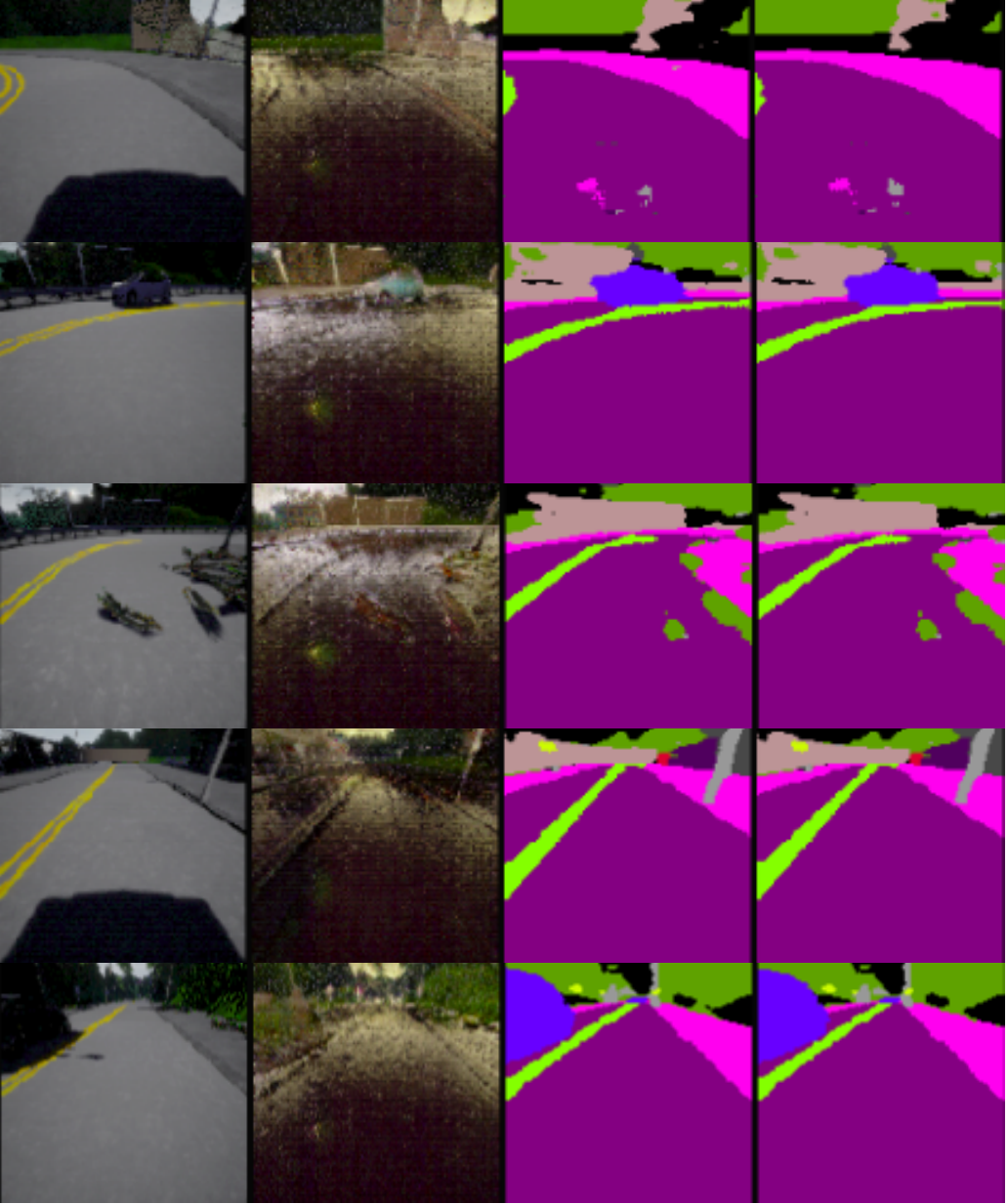}
  \caption{This figure is with reference to the master-servant architecture. \textbf{1\textsuperscript{st}~column:} The first column shows five sample images from domain $X$. \textbf{2\textsuperscript{nd}~column:} The second column shows corresponding images from domain $Y$, produced by the generator $G$, maintaining the semantics of the scene. \textbf{3\textsuperscript{rd}~column:} The segmentation reconstruction produced by feeding $z_1$ through the the master decoder. $z_1$ in turn is generated by feeding the images in the {2\textsuperscript{nd}~column} through the servant perception module $P_1$. \textbf{4\textsuperscript{th}~column:} Segmentation reconstruction produced by feeding $z_0$ through the master decoder. $z_0$ is generated by feeding the images in the {1\textsuperscript{st}~column} through the master perception module $P_0$. Note that the semantic reconstructions in the {3\textsuperscript{rd}~column} and {4\textsuperscript{th}~column} are almost indistinguishable.}
  \label{fig:masterservantSamples}
\end{figure}

\begin{figure}[ht]
  \centering
  \includegraphics[width=\linewidth]{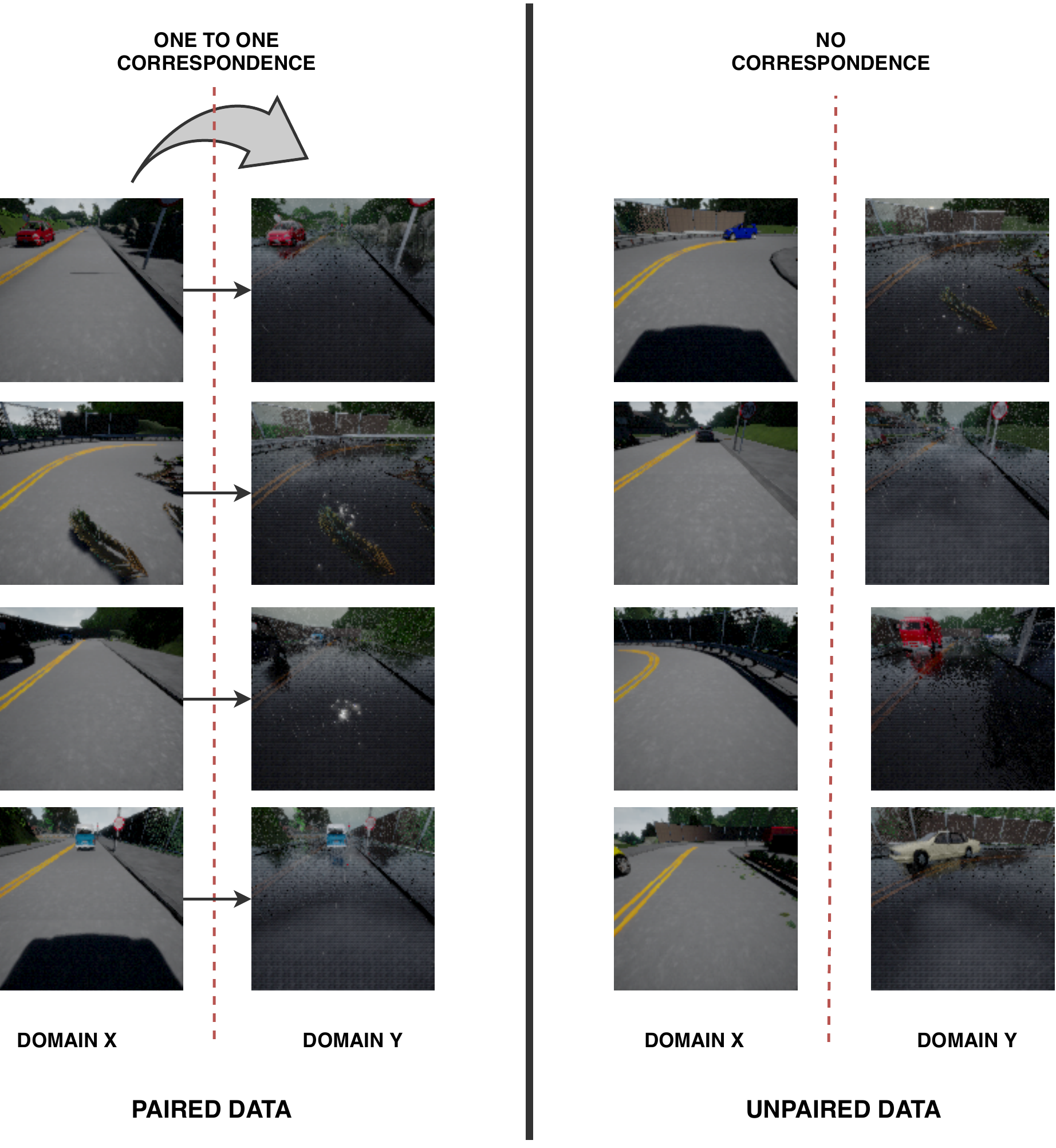}
  \caption{Examples of paired and unpaired dataset. Note that it is practically not possible to obtain an exact one-to-one correspondence between two differing road conditions. Hence, we use CycleGANs for image-to-image translation between unpaired images. The domains correspond to weather conditions of a sunny day and a rainy afternoon, respectively.} 
  \label{fig:pairedUnpaired}
\end{figure}

\begin{figure}[ht]
  \centering
  \includegraphics[width=\linewidth]{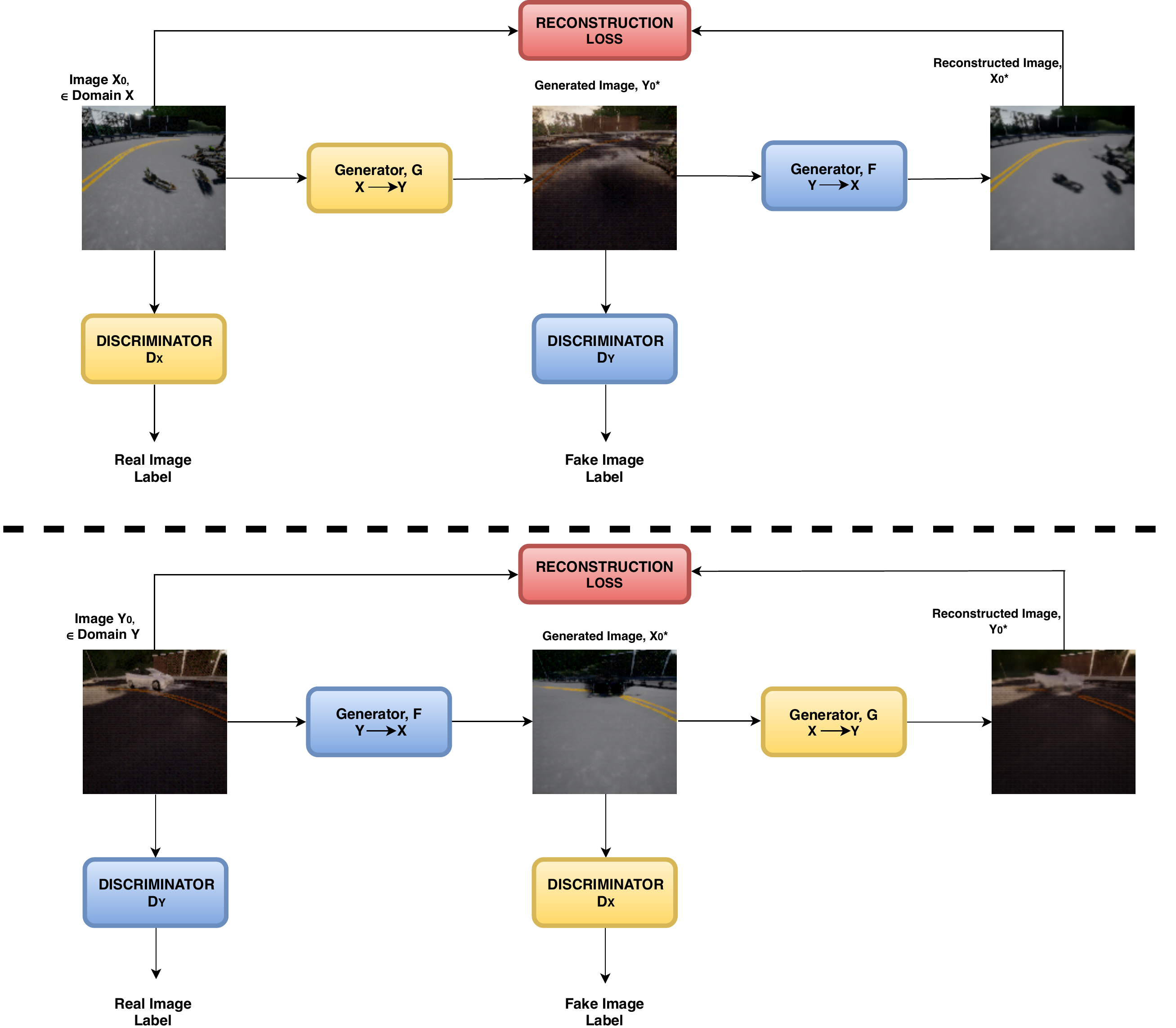}
  \caption{The critical component in the master-servant architecture in achieving unsupervised training of the servant perception module is the generator $G$, which transformed images from domain $X$ to domain $Y$, while maintaining the semantics of the scene. The generator $G$ is trained using CycleGANs. Unpaired images from domain $X$ and $Y$ produced by CARLA are used for training of the model. The top figure shows an arbitrary image $X_0$ from domain $X$ and is passed through the generator $G$, which generates an image $Y_0^*$. The generated image $Y_0^*$ is then fed to another generator $F$, which generates an image $X_0^*$. The network is optimized by minimizing the $L_1$ loss between the real image $X_0$ and the generated image $X_0^*$. To make the images appear realistic, each domain has its own discriminator network \ie $D_x$ and $D_y$. The bottom figure is analogous to the top one except that here, we fed a realistic image from domain $Y$ and try to minimize the $L_1$ loss between $Y_0$ and $Y_0^*$.}
  \label{fig:cgans}
\end{figure}

\begin{figure}[ht]
  \centering
  \includegraphics[width=0.6\linewidth]{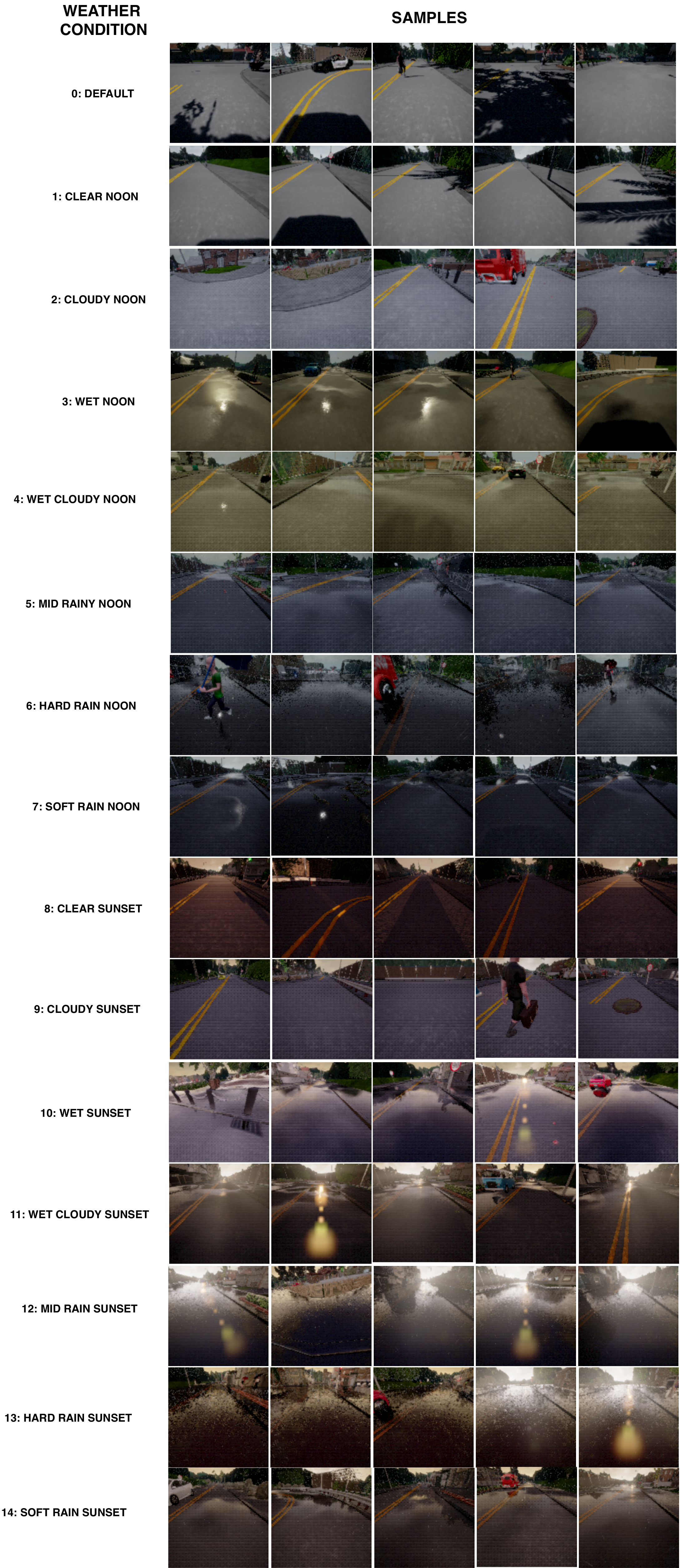}
  \caption{Some sample images of the 15 different weather conditions along with their description generated by the CARLA simulator. Note that some of the weather conditions are very similar and therefore, a perception module trained for one of the conditions may also work for a similar condition also.}
  \label{fig:carlaweathers}
\end{figure}

\begin{figure}[ht]
  \centering
  \includegraphics[width=\linewidth]{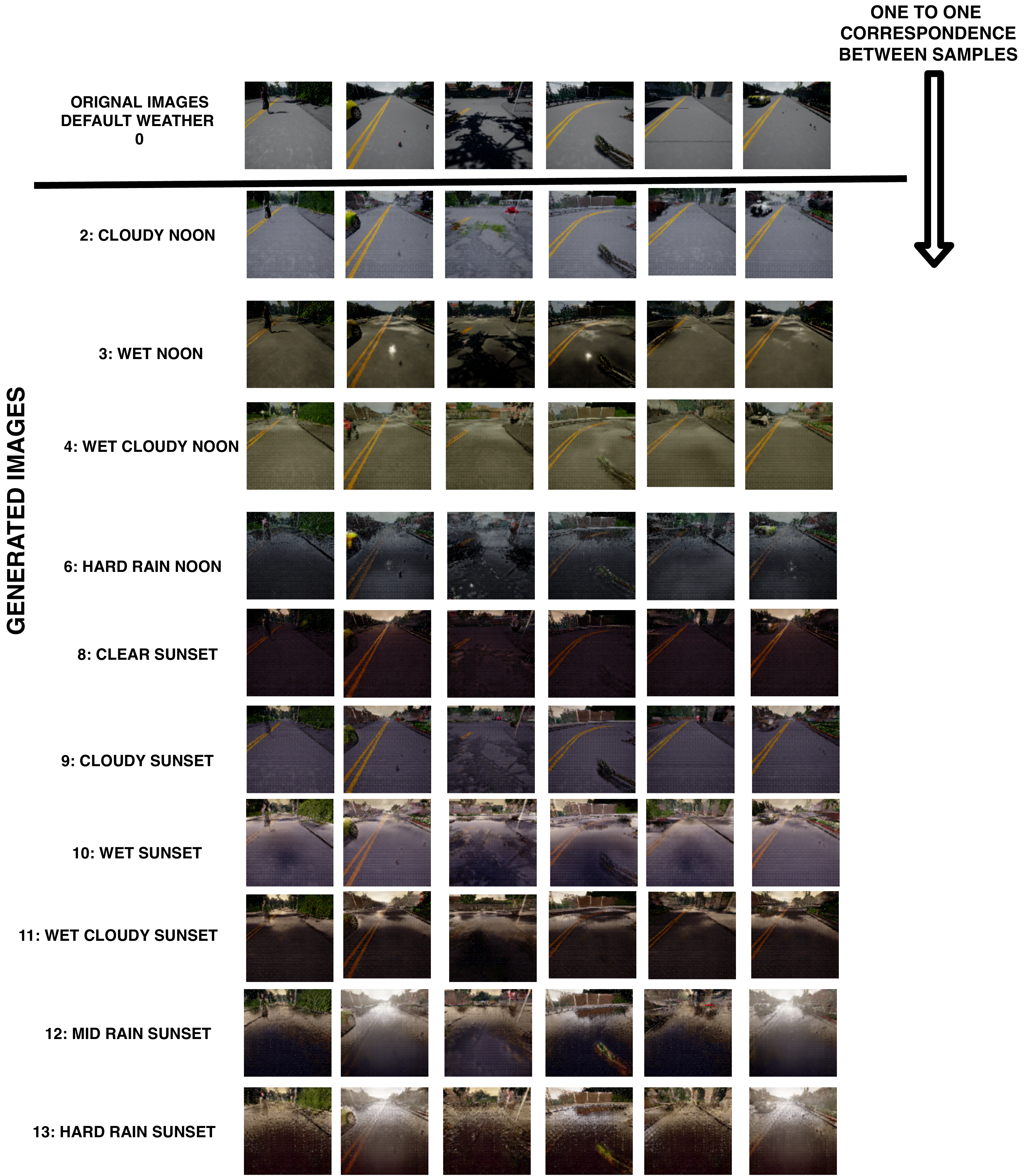}
  \caption{The figure shows 6 sample images generated from the original default condition for weather conditions 2, 3, 4, 6, 8, 9, 10, 11, 12, and 13 using the CycleGAN approach. The CyleGAN was trained with 3500 images from the default and each of the other weather conditions. Most of the generated images resemble the actual to a reasonable degree. For weather conditions with low visibility, \ie 12 and 13 some of the generated images (for \eg sample 3, 5, and 6) give a poor reconstruction.}
  \label{fig:cganexamples}
\end{figure}

\end{document}